\pdfoutput=1
\documentclass[11pt]{article}
\usepackage[a4paper,margin=2.4cm]{geometry}
\usepackage{amsmath,amssymb,booktabs,graphicx,microtype}
\usepackage{titlesec}
\titleformat{\section}{\normalfont\large\bfseries}{\thesection}{0.6em}{}
\titleformat{\subsection}{\normalfont\normalsize\bfseries}{\thesubsection}{0.6em}{}
\usepackage{hyperref}
\hypersetup{hidelinks,bookmarks=true}
\graphicspath{{./}}
\title{\vspace{-1.4cm}\textbf{What an odour descriptor corpus can and cannot
measure:\\ valence, attenuation, and the ceiling of the public record}}
\author{Stylianos Kampakis\thanks{Tesseract Academy, London, UK.
These authors contributed equally and are joint first authors.
Correspondence:
\texttt{stelios\string@thetesseractacademy.com},
\texttt{fabio@thetesseractacademy.com}.}
\and Fabio Rovai\footnotemark[1]}
\date{}

\begin{document}
\maketitle
\vspace{-0.8cm}

\begin{abstract}
\noindent
Machine olfaction trains on pooled public descriptor corpora. Whether a shared
descriptor word measures the same thing in two of them has not been tested, and
the prior question, of how much any of them can measure, has not been asked.

We audit four corpora distributed through Pyrfume. Conditioning on the molecule
makes McNemar's test the exact conditional test of the corpus effect, a
reduction we prove formally. Corpora disagree heterogeneously across descriptors
($I^2 = 80\%$) and non-uniformly with molecule labelling breadth ($z = 17.2$),
so no single offset repairs pooling. Separating criterion from construct, the
median tetrachoric agreement is 0.795 against a median $\kappa$ of 0.413: the
sources largely concur about which molecules deserve a word and differ about how
readily they apply it. Of 109 descriptors carrying an estimable effect, 36 show
large differential functioning on the ETS scale.

We then measure the ceiling. Against a human panel's own reliability, molecular
structure encoded as Morgan fingerprints with the full RDKit descriptor block
reaches 32.9\% of achievable, and adding every label in two merged corpora
reaches 33.9\%. Two thirds of reliably measurable perceptual similarity is in
neither representation, and the gap does not close with model capacity, with
three different molecular encodings, or with more molecules or more descriptor
words.

The missing variance is valence. One pleasantness rating per molecule reaches
54.6\% of achievable, replicated at 57.1\% from an independent instrument three
decades older. Valence is recoverable from descriptor labels at $\rho = 0.457$
and a coordinate built from that recovery reaches only 16.6\%, so it must be
measured rather than inferred. Five raters exceed structure plus the entire
descriptor record; fifteen to twenty saturate. We release a descriptor crosswalk
in which every term of both vocabularies appears with its evidence, and a
formal development of twenty machine-checked theorems.

\end{abstract}

\section{Introduction}

Public odour datasets are combined because they share molecules and appear to
share descriptor words. Neither overlap guarantees that the recorded variables
are commensurate. A descriptor may be elicited as a panel applicability
percentage, a slider rating, a controlled binary label, or a term extracted from
a commercial record, and identical spelling need not imply an identical response
process.

The vocabularies themselves are a first indication that the problem is real. Of
the 666 descriptor terms GoodScents applies and the 113 Leffingwell applies, 111
match exactly after normalisation, so a merge of the two either discards or
invents 555 terms of one vocabulary. Counting overlap does not settle whether
the 111 that do match mean the same thing.

This paper asks what follows for measurement: whether the disagreement between
sources is one of criterion or of construct, what a pooled corpus can be
corrected to support, and what the record can measure at all. It is
self-contained: every number reported below is computed here from the pinned
sources named in Section 2.

Our contribution is in four parts. We give the exact conditional analysis this
design admits and prove why it is exact. We separate criterion from construct
and report the result as differential item functioning with equating constants.
We measure how much of reliably measurable human perceptual similarity the
record's representations account for, and find it is about a third. And we
identify the missing majority as valence, defend that identification against
five distinct objections, and price the measurement that would supply it.

\begin{figure}[t]
\centering
\includegraphics[width=0.84\textwidth]{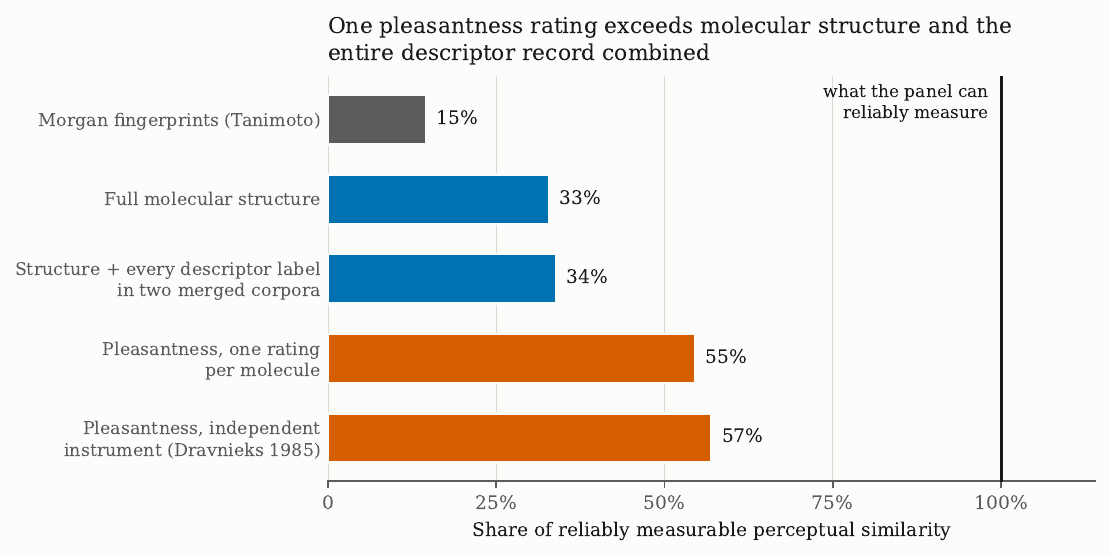}
\caption{The ceiling of the public record, and what fills it. Performance is
expressed as a share of what the panel criterion's own reliability permits, so
that a correlation against a noisy target can be read. Molecular structure and
every descriptor label in two merged corpora reach about a third. One
pleasantness rating per molecule reaches more than half, and replicates from an
instrument three decades older.}
\label{fig:headline}
\end{figure}

\section{Materials and methods}

\subsection{Corpora}

We used Dravnieks, Keller and Vosshall, Leffingwell and GoodScents as
distributed in Pyrfume, pinned to a single commit and verified against a
manifest of byte counts and SHA-256 digests. Molecules are keyed on positive
PubChem CID; records with missing, zero or negative identifiers are excluded.
Two Keller stimulus molecules carry a CAS registry number in the identifier
field and are resolved through the accompanying molecule table rather than
dropped, which accounts for a discrepancy of two molecules in anchor counts
reported elsewhere. Descriptor strings are stripped, lower-cased, and every run
of whitespace, underscore or hyphen collapsed to one space.

\begin{center}
\small
\begin{tabular}{@{}lrrl@{}}
\toprule
\textbf{Corpus} & \textbf{Molecules} & \textbf{Descriptors} & \textbf{Response generation}\\
\midrule
Dravnieks (1985) & 138 & 146 & panel applicability percentage\\
Keller et al.\ (2016) & 480 & 20 & per-subject slider, 55 subjects\\
Leffingwell & 3{,}487 & 113 & controlled vocabulary attribution\\
GoodScents & 4{,}565 & 666 & free-text derived attribution\\
\bottomrule
\end{tabular}
\end{center}

Leffingwell and GoodScents share 2{,}225 molecules and 111 exactly matching
normalised descriptors, and are the only well-powered pair. Both are trade
compilations and contain no human observer, a limitation we address in
Section~\ref{sec:ceiling} by importing a criterion that does.

\subsection{The exact conditional analysis}

Each molecule-descriptor cell carries exactly two observations, one per corpus.
Writing the labelling process as a crossed logistic model with a molecule
effect, a molecule-by-descriptor effect and a descriptor-specific corpus effect,
conditioning on the pair total removes the first two as nuisance parameters
without any distributional assumption. The sufficient statistic is the
discordant-pair count and the exact conditional test is McNemar's. We prove this
formally: for a nuisance term $\mu$ and corpus effect $\delta$,
\[
\frac{\Pr(\text{B alone})}{\Pr(\text{A alone}) + \Pr(\text{B alone})}
  = \sigma(\delta),
\]
with $\mu$ cancelling exactly, and the conditional maximum-likelihood estimate
of $\delta$ is the discordant log odds ratio. Both statements are machine-checked
in Lean 4 against Mathlib, with no axioms beyond \texttt{propext},
\texttt{Classical.choice} and \texttt{Quot.sound}.

Descriptors whose discordant cells contain a zero have an unbounded conditional
estimate and are reported separately rather than pooled.

\subsection{Criterion and construct}

Conditioning removes exactly the effects needed to ask whether the corpora agree
about \emph{which} molecules deserve a word, so a second estimand is required.
Per descriptor, the paired table has three free cell proportions and a saturated
bivariate probit spends exactly three parameters on it, two thresholds and one
tetrachoric correlation. It is exactly identified with zero degrees of freedom
and assumes nothing about dimensionality, local independence, a latent molecule
trait, or a set of unaffected anchor items. The thresholds are the criterion
quantity and the correlation is the construct quantity.

\subsection{The perceptual criterion}
\label{sec:criterion}

Keller and Vosshall recorded 55 subjects rating molecules individually, so a
molecule's mean panel profile is a description given by people and the
correlation between two profiles is a perceptual similarity that no curator
produced. Its reliability is estimated from the panel's own split half and
corrected to full-panel size, which caps any predictor's correlation at the
square root of that reliability. We therefore report performance as a share of
the achievable rather than as a bare number.

Molecular structure is encoded as Morgan fingerprints of radius 2 with 2048
bits, giving a Tanimoto similarity, together with the full RDKit physicochemical
descriptor block, of which 180 columns vary. Splits are grouped by molecule and
a held-out pair must have both of its molecules held out, because a pair sharing
one molecule with the training set leaks.

\section{Results}

\subsection{Corpora disagree heterogeneously and non-uniformly}

Across the 2{,}225 shared molecules, mean per-molecule descriptor-set Jaccard
agreement is 0.359 and only 5.3\% of molecules carry identical positive
descriptor sets. Identical identifiers and identical vocabulary do not imply
identical labels.

Pooling the descriptor-level conditional effects, the average corpus difference
is negligible while the descriptor-level differences are not. The pooled odds
ratio is 1.14 with a 95\% interval of 1.02 to 1.28, and Cochran's $Q$ is 528 on
108 degrees of freedom. Between-descriptor heterogeneity is $I^2 = 80\%$ with
$\tau = 0.52$, and the prediction interval for the next descriptor spans odds of
0.39 to 3.21. No single offset repairs pooling.

The effect is also non-uniform. Matching on molecule labelling breadth, the
conditional log odds ratio runs from $-0.511$ for molecules carrying one to four
descriptors to $+0.485$ for those carrying seven to fifteen, a trend of $+0.390$
per stratum at $z = 17.2$. The direction of the corpus effect turns with how
much is said about the molecule, so a per-descriptor offset is insufficient as
well.

\begin{figure}[t]
\centering
\includegraphics[width=0.86\textwidth]{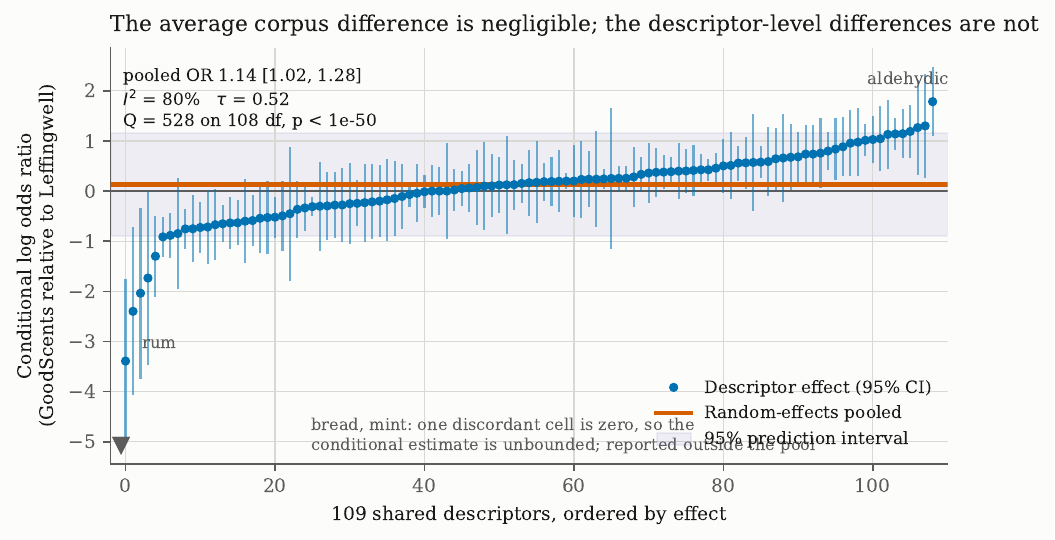}
\caption{Descriptor-level conditional effects with the random-effects pooled
estimate and prediction interval. The average difference is negligible; the
descriptor-level differences are not.}
\end{figure}

\subsection{The disagreement is criterion, not construct}

Median tetrachoric agreement across descriptors is 0.795 with a bootstrap
interval of 0.768 to 0.821, against a median Cohen's $\kappa$ of 0.413, a gap of
0.382. Ninety-three per cent of descriptors exceed 0.5. The corpora largely
concur about which molecules deserve a descriptor and differ heterogeneously
about how readily they apply it.

We state before a reader computes it that $\rho$ ranks descriptors almost
exactly as $\kappa$ does, at Pearson and Spearman 0.865. What $\rho$ supplies is
the level, not the ordering, and the claim is that observed agreement understates
concurrence by roughly 0.38.

\subsection{Differential item functioning and equating}

The conditional log odds ratio is the Mantel-Haenszel statistic under
stratification on the molecule, which is the finest matching available, and
under matched pairs that estimator reduces to McNemar's exactly. Reported on the
ETS delta scale over 109 estimable descriptors, 56 show negligible differential
functioning, 17 intermediate and 36 large. The largest are \emph{rum},
\emph{gasoline}, \emph{chamomile}, \emph{aldehydic} and \emph{catty}.

The probit thresholds are equating constants, since both latent propensities are
standard normal by construction and the transformation between criteria is a
shift. We release all 110 with bootstrap standard errors; the median absolute
constant is 0.102 and the range runs from $-0.512$ to $+1.515$. Item-response
equating, meaning Stocking-Lord or Haebara, is not performed and not claimed,
because those methods link fitted item-response models and a unidimensional one
is misspecified for these descriptors.

\subsection{The ceiling of the public record}
\label{sec:ceiling}

Against the panel criterion of Section~\ref{sec:criterion}:

Figure~\ref{fig:headline} summarises this section and the next.

\begin{center}
\small
\begin{tabular}{@{}lrr@{}}
\toprule
\textbf{Representation} & \textbf{$\rho$} & \textbf{Share of achievable}\\
\midrule
Tanimoto on Morgan fingerprints alone & 0.131 & 14.6\%\\
Full RDKit structural block & 0.296 & 32.9\%\\
Plus every label in two merged corpora & 0.306 & \textbf{33.9\%}\\
\bottomrule
\end{tabular}
\end{center}

Two thirds of reliably measurable perceptual similarity is explained by neither
molecular structure nor the descriptor record. The gap does not close three
independent ways. It is not model capacity, since a gradient booster matches a
ridge to within 0.007. It is not representation, since a twenty-feature
encoding, a character n-gram embedding and the full RDKit block all land between
30 and 34\%. And it is not volume, since the descriptor contribution is flat in
the number of molecules labelled and in vocabulary size.

\subsection{The missing variance is valence}

Keller records three items excluded from every analysis above as non-character
responses: how pleasant, how strong, how familiar. Trade corpora record none of
them. Splitting the 55 subjects so that one half supplies the affective items
and the other supplies the criterion removes the shared-rater confound.

\begin{center}
\small
\begin{tabular}{@{}lrr@{}}
\toprule
\textbf{Predictors} & \textbf{$\rho$} & \textbf{Share of achievable}\\
\midrule
Structure and every descriptor label & 0.303 & 33.6\%\\
Pleasantness alone & \textbf{0.492} & \textbf{54.6\%}\\
Intensity alone & 0.130 & 14.5\%\\
Familiarity alone & 0.105 & 11.7\%\\
\bottomrule
\end{tabular}
\end{center}

One number per molecule outperforms molecular structure and two merged corpora
combined. Dropping pleasantness collapses the affective set from 0.504 to 0.148
while dropping intensity or familiarity moves it by under 0.01, so the axis is
valence rather than arousal or exposure.

\begin{figure}[t]
\centering
\includegraphics[width=0.80\textwidth]{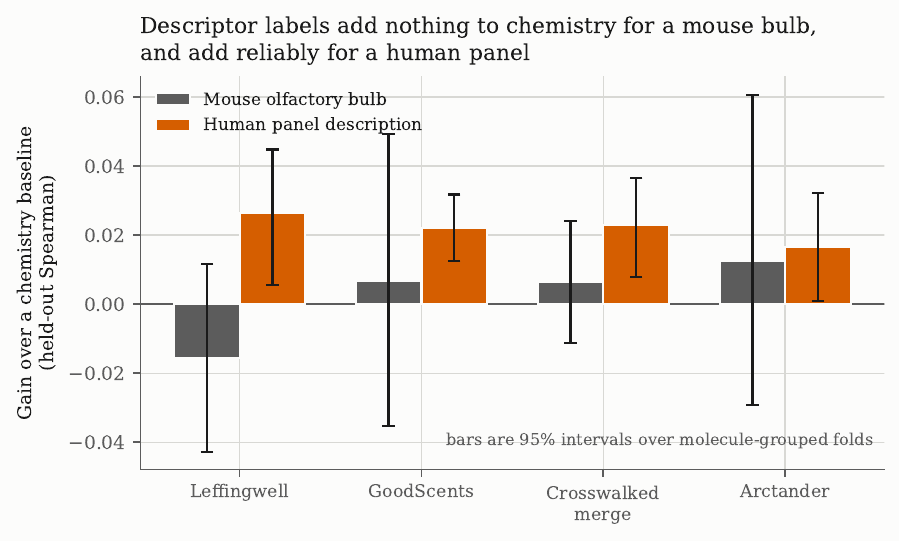}
\caption{Descriptor labels add nothing to chemistry against a mouse olfactory
bulb, and add reliably against a human panel. Bars are 95\% intervals over
molecule-grouped folds.}
\end{figure}

\subsection{Five objections}

\textbf{Shared method.} Pleasantness and the criterion share an instrument.
Dravnieks carries explicitly evaluative descriptors, \emph{fragrant} against
\emph{sickening}, \emph{putrid}, \emph{rancid} and \emph{fecal}, from a
different panel, decade and response format. On the 64 molecules it shares with
Keller and the corpora, that proxy alone reaches 57.1\% of achievable against
28.3\% for structure plus every descriptor label. One feature from a 1985 atlas
exceeds 183 features of modern cheminformatics and merged trade vocabulary.

\textbf{Recovery from labels.} Valence might be implicit in quality words. It is
partly recoverable and the recovery is insufficient: corpus labels predict
pleasantness at $\rho = 0.457$ out of fold, but a valence coordinate built from
that prediction reaches 16.6\% of achievable against 53.9\% for measured valence
on the same molecules and folds. This is the recoverability boundary of
Section~\ref{sec:formal} in empirical form.

\textbf{Multiplicity.} Against a null shuffling molecule identity while
preserving the valence marginal, the observed 0.486 exceeds every one of 200
shuffles, whose mean is $+0.002$ and maximum $+0.048$.

\textbf{Model and representation} are addressed in Section~\ref{sec:ceiling}.

\subsection{How much measurement is needed}

\begin{center}
\small
\begin{tabular}{@{}rrrrrrr@{}}
\toprule
Raters & 1 & 3 & \textbf{5} & 10 & 20 & 27\\
Share of achievable & 16.4\% & 25.9\% & \textbf{39.2\%} & 45.6\% & 52.2\% & 53.9\%\\
\bottomrule
\end{tabular}
\end{center}

Five raters exceed structure plus every descriptor label. Twenty to twenty-seven
buys 1.7 points, so fifteen to twenty is the practical point. This is the only
axis on which collection buys anything, and it buys it quickly and then stops.

\subsection{A correction to our own dimensionality evidence}

We first reported a leading factor of 3.2\% of item variance as evidence that no
dominant dimension organises these descriptors. That figure comes from Pearson
correlations between sparse binary labels, which this paper elsewhere shows
understate association by roughly 0.38. On the tetrachoric matrix the leading
factor is 21.2\% of trace, and it is valence: its molecule scores correlate with
measured pleasantness at $-0.473$, loading negatively on \emph{apple},
\emph{floral}, \emph{honey} and \emph{rose} and positively on \emph{coffee},
\emph{savory}, \emph{onion} and \emph{sulfurous}.

The refusal of a unidimensional item-response model stands, since the first to
second eigenvalue ratio is 1.22 and 53 of 111 eigenvalues exceed one. What does
not stand is the stronger reading that no interpretable dominant dimension
exists. One does, and our own attenuation result explains why a naive analysis
misses it.

\subsection{Confirmation on an untouched pair}

A specification was written and committed before execution, naming
\texttt{ifra\_2019} against \texttt{sharma\_2021a}, a pair used nowhere in
developing any result above, with four numbered predictions. The analysis was
run once on 834 shared molecules and 158 shared descriptors.

Heterogeneity confirmed at $I^2 = 64.1\%$ with $Q = 317$ on 114 degrees of
freedom. The attenuation gap confirmed at $+0.463$. Construct agreement
confirmed at a median tetrachoric of 0.835. The prediction that the average
corpus effect is small \textbf{failed}, at $+1.426$ in log odds against a
threshold of 0.5, and is reported as failed. The exploratory finding that
corpora agree on average is a property of the pairs examined, not of corpus
pairs in general, and pooling must be corrected for a large systematic offset as
well as for the descriptor-level spread that no offset removes.

\section{Formal results}
\label{sec:formal}

Twenty theorems are machine-checked in Lean 4 against Mathlib, with no
\texttt{sorry} and no axioms beyond the three standard ones.

\textbf{Conditioning.} The nuisance term cancels exactly, so McNemar's test is
the exact conditional test of the corpus effect and the discordant log odds
ratio is its conditional maximum-likelihood estimate.

\textbf{Recoverability.} A crosswalk recovers a target from a recording if and
only if the recording never merges two states the target must distinguish, with
the corollary that exhibiting one merged pair refutes every possible crosswalk
at once. This separates a calibration failure that a different rule can undo
from an information failure that nothing can undo.

\textbf{Capacity.} A recording made of $k$ binary descriptors takes at most
$2^k$ values, so a target needing more distinctions is unreachable by any
crosswalk. The bound does not mention raters.

\textbf{Aggregation.} A three-rater majority applies a label with probability
$3p^2 - 2p^3$, which is below $p$ when $p < \tfrac12$ and above it when
$p > \tfrac12$. Read with $p$ as accuracy this is the Condorcet jury theorem, so
more raters do help; read with $p$ as an application rate under a no-signal
model it is prevalence shrinkage, which a different threshold undoes. Limits
under thresholding depend on the regime: a fixed vote requirement drives
agreement to one, unanimity drives it to zero.

\section{Discussion}

The public descriptor record does not carry the dominant dimension of odour
perception, and cannot be made to yield it by inference. That is a statement
about what to collect rather than about how to model. More compounds do not
close the gap and more quality words do not close it, and both were tested. Five
people rating pleasantness do.

For a resource intending to support benchmarking, three consequences follow.
Pooling requires correction for a systematic offset and a heterogeneous
descriptor-level component, and the non-uniformity means neither is a constant.
A crosswalk is a versioned object whose match types are claims requiring
evidence, and we release one in which all 668 terms of both vocabularies appear,
111 matched and 557 retained as unmatched so that lost coverage is visible.
And a valence coordinate should be collected at source rather than derived.

\subsection{Limitations}

The main effect rests on one criterion dataset, with an external replication on
64 molecules. The criterion is a panel's descriptor profile, so what is
predicted is how people apply words rather than percept unmediated by language,
and some of the unexplained share may be word choice. Valence being the leading
axis of odour perception is established in the psychophysics literature; the
contribution here is quantifying the record's failure to carry it and pricing
the remedy, not the discovery of the axis. PubChem CID links chemical identity
but not stereochemistry, concentration, solvent or purity. The Dravnieks
concentration field is not a measurement: it is reproduced exactly, for all 160
stimulus rows, by testing whether the stimulus name contains the substring
\texttt{low}, and only six rows carry that value. Analyses that aggregate
Dravnieks across concentrations inherit this. The two trade corpora are not established as independent, since
both draw on the same primary literature.

A downstream demonstration could not be run. Only four descriptors are carried
by both corpora and an independent human panel often enough to score, and the
differences between training regimes are around 0.005 in AUC. That is a
statement about the record rather than a null result, and it joins two others:
the only two human descriptor panels share two usable descriptors, and no human
perceptual criterion in the public record is large enough to validate labels
against. The record cannot presently validate its own labels.

\section*{Data and code availability}

Corpora and their reuse positions are listed in the methods; the source snapshot
is identified by a pinned commit so inputs can be retrieved independently, and
every file was checked against a manifest of byte counts and digests. Raw corpus
rows are not redistributed, because the reuse positions do not permit it. The
analysis code and derived reports are not publicly released. An appendix gives
every procedure in sufficient detail to reimplement, including seven pitfalls
that produced incorrect results during development.

\end{document}